\documentclass[sigconf,review=false,anonymous=false,authordraft=false]{acmart}

\usepackage{arydshln}
\usepackage{mathbbol}
\usepackage{amsmath}
\usepackage{nicematrix}

\copyrightyear{2024}
\acmYear{2024}
\setcopyright{rightsretained}

\acmConference[SIGIR '24]{Proceedings of the 47th International ACM SIGIR Conference on Research and Development in Information Retrieval}{July 14--18, 2024}{Washington, DC, USA.}
\acmBooktitle{Proceedings of the 47th Int'l ACM SIGIR Conference on Research and Development in Information Retrieval (SIGIR '24), July 14--18, 2024, Washington, DC, USA}

\makeatother

\usepackage{enumitem}
\newlist{inlinelist}{enumerate*}{1}
\setlist*[inlinelist,1]{%
  label=(\roman*),
}
\usepackage{subfigure}
\usepackage{multirow}
\usepackage{adjustbox}
\usepackage{xspace}

\newcommand{\metric}{eRAG\xspace}

\title{Evaluating Retrieval Quality in Retrieval-Augmented Generation}

\author{Alireza Salemi}
\affiliation{\institution{University of Massachusetts Amherst}
\city{Amherst}
\state{MA}
\country{United States}}
\email{asalemi@cs.umass.edu}

\author{Hamed Zamani}
\affiliation{\institution{University of Massachusetts Amherst}
\city{Amherst}
\state{MA}
\country{United States}}
\email{zamani@cs.umass.edu}

\begin{document}


\begin{abstract}

Evaluating retrieval-augmented generation (RAG) presents challenges, particularly for retrieval models within these systems. Traditional end-to-end evaluation methods are computationally expensive. Furthermore, evaluation of the retrieval model's performance based on query-document relevance labels shows a small correlation with the RAG system's downstream performance. We propose a novel evaluation approach, \metric, where each document in the retrieval list is individually utilized by the large language model within the RAG system. The output generated for each document is then evaluated based on the downstream task ground truth labels. In this manner, the downstream performance for each document serves as its relevance label. We employ various downstream task metrics to obtain document-level annotations and aggregate them using set-based or ranking metrics. Extensive experiments on a wide range of datasets demonstrate that \metric achieves a higher correlation with downstream RAG performance compared to baseline methods, with improvements in Kendall's $\tau$ correlation ranging from 0.168 to 0.494. Additionally, \metric offers significant computational advantages, improving runtime and consuming up to 50 times less GPU memory than end-to-end evaluation.
\end{abstract}




\maketitle

\section{Introduction}

Retrieval-augmented generation (RAG) has emerged as a prominent approach in natural language processing, combining the strengths of retrieval and generation models \cite{reml}, with use cases in decreasing hallucination \cite{agrawal2023knowledge, shuster-etal-2021-retrieval-augmentation}, knowledge-grounding \cite{rag,fid,srag}, and personalization \cite{salemi2023lamp, salemi2024optimization}. Evaluating RAG systems is important as it ensures the effectiveness of integrating retrieval-based methods with generative models \cite{ares,ragas}. Traditionally, RAG evaluation has primarily relied on end-to-end assessment, which entails comparing the generated output with one or more ground truth references \cite{kilt}. While this is crucial, it presents several limitations, especially, for evaluating retrieval models in RAG systems.

First, end-to-end evaluation lacks transparency regarding which retrieved document contributed to the generated output, hindering interpretability of the system's behavior. Secondly, it is resource-intensive, consuming significant time and computational power, particularly when dealing with a large set of retrieval results consumed by the LLM. To process long input sequences resulting from the utilization of all retrieved documents by the LLM, GPUs with substantial memory capacities are essential for end-to-end evaluation. Moreover, many ranking systems rely on interleaving (i.e., replacing one or more documents in the result list) for evaluation and optimization, which further complicates the evaluation, as slight variations in retrieval results necessitate re-computation of the RAG pipeline. Finally, optimizing ranking models often requires document-level feedback, such as user clicks \cite{10.1145/3539618.3591639, 10.1145/1498759.1498818}. However, end-to-end evaluation only provides list-level feedback for the retrieval results. That said, this paper studies retrieval evaluation in RAG.

Human annotations can be a potential solution for evaluating retrieval models in RAG, however, accurate annotations are often challenging and costly to obtain. More recently, with the emergence of large language models (LLMs) and their advanced capabilities in reasoning and text comprehension, they have been utilized to annotate documents for retrieval evaluation \cite{ares, ragas}. Nevertheless, these approaches predominantly evaluate the retriever in RAG systems based on human preferences, whereas the primary objective of the retrieval model in RAG is to serve the LLM that leverages the retrieved results \cite{reml}. That said, our extensive investigation on a diverse set of RAG systems for open-domain question answering, fact verification, and dialogue systems reveals that employing human annotations, such as the \textit{provenance} labels in the KILT benchmark \cite{kilt}, for evaluating the retrieval models within a RAG system exhibits only a minor correlation with the downstream RAG performance. This indicates a lack of meaningful relationship between the evaluated metrics and the downstream performance of RAG.


In this paper, we propose \metric, a new approach for \underline{e}valuating retrievers in \underline{RAG} systems, where we apply the LLM in RAG system on each document in the retrieval result list individually and use the LLM's output to provide document-level annotations. These annotations can be obtained using any arbitrary downstream task metric, such as accuracy, exact match, or ROUGE \cite{lin-2004-rouge}. We can then apply a set-based or ranking metric as an aggregation function to obtain a single evaluation score for each retrieval result list. 

We evaluate our proposed approach on question answering, fact-checking, and dialogue generation from the knowledge-intensive language tasks (KILT) benchmark \cite{kilt}. Our results demonstrate that our proposed approach achieves the highest correlation with the downstream performance of the RAG system in comparison with the baselines. Specifically, we observe an absolute improvement in Kendall's tau correlation ranging between 0.168 and 0.494 across the evaluated datasets. Furthermore, we investigate the impact of different retrieval augmentation methods, the quantity of retrieved documents, and the LLM size on correlation. Finally, we demonstrate that our approach offers significant computational advantages, consuming up to 50 times less memory compared to end-to-end evaluation. To facilitate research in this domain, we make eRAG's implementation publicly available at: {\url{https://github.com/alirezasalemi7/eRAG}}.

\section{Evaluating Retrievers in RAG}

Generally, two predominant methods are used for obtaining relevance labels for retrieval evaluation. The first approach involves human judgment to assess the relevance of a query to documents within a corpus. The main issue with this approach is that human annotation can be costly and is often impractical for evaluating all documents in a corpus \cite{scott-etal-2012-corpus}. Moreover, human annotation relies on human preferences to judge the relevance of documents to a query. However, a document deemed relevant based on human preferences may not be useful for an LLM in fulfilling its task. 

The second approach utilizes the downstream ground truth output associated with the query to provide weak relevance labels. In this method, a retrieved document containing the downstream ground truth is considered relevant \cite{dpr, izacard2021distilling, 10.1145/3578337.3605137, 10.1145/3539618.3591629}. This method also presents its own challenges. This approach is impractical, particularly in scenarios where the task involves long-text generation or text classification, as downstream task labels might not exist within documents. Also, one document can be useful for an LLM in fulfilling its task without containing the ground truth labels.

Even though we are not aware any work that use LLMs for evaluating retrieval models in RAG, LLMs can be leveraged to label documents based on their relevance to a query. Inspired by \citet{thomas2023large}, the LLM functions as a binary classifier, indicating whether a document is relevant to the query or not. The mentioned challenges persist even with the judgment of LLMs, especially if the LLM responsible for labeling differs from the LLM in the RAG pipeline. Besides, employing LLMs as judges in this scenario can pose challenges due to the computational cost of running them on a large set of retrieved documents and memory constraints.

To mitigate these problems, we propose \metric, a novel approach that involves utilizing the LLM in RAG system itself as the arbiter for generating labels to evaluate the retrieval model.

\subsubsection*{\textbf{Using Downstream Large Language Model in RAG as Document Annotator}}

Consider a retrieval model $\mathcal{R}$ that produces a ranked list $\mathbf{R}_k$ with $k$ documents for the LLM $\mathcal{M}$ tasked with performing a specific task, utilizing a downstream evaluation function $\mathcal{E}_{\mathcal{M}}$. The LLM $\mathcal{M}$ takes a ranked list of documents as its input along with the query $q$, and generates an output represented as $\bar{y} = \mathcal{M}(q, \mathbf{R}_k)$. 
For the documents in $\mathbf{R}_k$, we feed each document individually to the LLM $\mathcal{M}$ with the query and evaluate the generated answer to create the label for each document, expressed as:

\begin{equation}
    {\mathcal{G}_q}[d] = \mathcal{E}_{\mathcal{M}}(\mathcal{M}(q, \{d\}), y) \quad : \quad \forall d \in \mathbf{R}_k
\end{equation}

\noindent
where $y$ is the expected downstream output for the query. We can employ the created $\mathcal{G}_q$ to utilize any ranking metric to evaluate $\mathcal{R}$.

Note that the runtime cost of a vanilla transformer \cite{NIPS2017_3f5ee243} scales quadratically with its input length. Consequently, for end-to-end evaluation, the cost of running a transformer on a ranked list with $k$ documents, with an average length of $d$, to generate an output with length $l$ is $O(lk^2d^2)$. Conversely, in our approach, as each document is individually fed to the LLM for k times, the cost is $O(lkd^2)$, proving to be more efficient than end-to-end evaluation.

\subsubsection*{\textbf{Retrieval Evaluation Metrics}}

For a ranked list $\mathbf{R}_k$, comprising $k$ retrieved documents generated by a retrieval model $\mathcal{R}$, an evaluation metric $\mathcal{E}_\mathcal{R}$ assigns a score ${\mathcal{E}_\mathcal{R}(\mathbf{R}_k, \mathcal{G}_q)} \in [0, 1]$, by comparing the ranked list with the relevance scores $\mathcal{G}_q$, which is a function that maps each document to a scalar relevance score for the document with respect to the query $q$ (i.e., $\mathcal{G}_q(d) = s_d$). Various definitions exist for the evaluation metric $\mathcal{E}_\mathcal{R}$; in this paper, we examine Precision (P), Recall (R), Mean Average Precision (MAP), Mean Reciprocal Rank (MRR) \cite{mrr}, Normalized Discounted Cumulative Gain (NDCG) \cite{ndcg}, and Hit Rate. Note that when dealing with non-binary relevance labels, precision considers the average value of relevance labels, while Hit Ratio considers the maximum value among them. 

\section{Experiments}

\begin{table*}[]
    \centering
    \caption{The correlation between each evaluation approach and the downstream performance of the LLM. T5-small with FiD with 50 retrieved documents is used. We do not report correlation for the Answers method for FEVER and WOW datasets because the answers to queries do not exist in the document since FEVER is a classification dataset and WoW is long-text generation. For the WoW dataset, we only report correlation on Precision and Hit Ratio because other metrics do not support non-integer relevance labels. Tau is Kendall's tau and rho is Spearman's rho.}
    \vspace{-0.4cm}
    \adjustbox{max width=\textwidth}{
    \begin{NiceTabular}{c|c|cc|cc|cc|cc|cc|cc|cc|cc|cc|cc}
         \multirow{3}{*}{\shortstack[c]{Relevance \\ Annotation}} & \multirow{3}{*}{Metric} & \multicolumn{10}{c|}{BM25} & \multicolumn{10}{|c}{Contriever} \\
         \cline{3-22}
         & & \multicolumn{2}{c}{NQ} & \multicolumn{2}{c}{TriviaQA} & \multicolumn{2}{c}{HotpotQA} & \multicolumn{2}{c}{FEVER} & \multicolumn{2}{c|}{WoW} & \multicolumn{2}{|c}{NQ} & \multicolumn{2}{c}{TriviaQA} & \multicolumn{2}{c}{HotpotQA} & \multicolumn{2}{c}{FEVER} & \multicolumn{2}{c}{WoW} \\
         \cline{3-22}
         & & tau & rho & tau & rho & tau & rho & tau & rho & tau & rho & tau & rho & tau & rho & tau & rho & tau & rho & tau & rho \\

         \hline
         \multirow{6}{*}{\shortstack[c]{Containing \\ the \\ Answer}} & MAP & 0.349 & 0.417 & 0.298 & 0.364 & 0.359 & 0.423 & - & - & - & - & 0.303 & 0.366 & 0.265 & 0.325 & 0.379 & 0.429 & - & - & - & - \\
         & MRR & 0.361 & 0.417 & 0.313 & 0.340 & 0.398 & 0.449 & - & - & - & - & 0.301 & 0.353 & 0.257 & 0.292 & 0.384 & 0.430 & - & - & - & - \\
         & NDCG & 0.357 & 0.427 & 0.298 & 0.365 & 0.370 & 0.435 & - & - & - & - & 0.313 & 0.378 & 0.270 & 0.331 & 0.385 & 0.437 & - & - & - & - \\
         & P & 0.353 & 0.411 & 0.276 & 0.333 & 0.396 & 0.454 & - & - & - & - & 0.346 & 0.403 & 0.283 & 0.340 & 0.406 & 0.449 & - & - & - & - \\
         & R & 0.325 & 0.325 & 0.232 & 0.232 & 0.375 & 0.375 & - & - & - & - & 0.319 & 0.319 & 0.215 & 0.215 & 0.401 & 0.401 & - & - & - & - \\
         & Hit Ratio & 0.325 & 0.325 & 0.232 & 0.232 & 0.375 & 0.375 & - & - & - & - & 0.319 & 0.319 & 0.215 & 0.215 & 0.401 & 0.401 & - & - & - & - \\

         \hline
         \multirow{6}{*}{\shortstack[c]{KILT \\ Provenance}} & MAP & 0.181 & 0.218 & 0.142 & 0.172 & 0.007 & 0.009 & 0.026 & 0.032 & 0.015 & 0.021 & 0.161 & 0.196 & 0.113 & 0.137 & 0.128 & 0.155 & 0.045 & 0.056 & 0.055 & 0.080 \\
         & MRR & 0.177 & 0.205 & 0.151 & 0.175 & 0.074 & 0.080 & 0.036 & 0.040 & 0.013 & 0.017  & 0.152 & 0.173 & 0.120 & 0.136 & 0.151 & 0.169 & 0.045 & 0.049 & 0.059 & 0.081 \\
         & NDCG & 0.179 & 0.216 & 0.142 & 0.172 & 0.021 & 0.026 & 0.029 & 0.036 & 0.013 & 0.019 & 0.159 & 0.193 & 0.115 & 0.140 & 0.134 & 0.162 & 0.045 & 0.056 & 0.056 & 0.081 \\
         & P & 0.163 & 0.192 & 0.140 & 0.165 & 0.139 & 0.164 & 0.043 & 0.051 & 0.011 & 0.015 & 0.131 & 0.157 & 0.108 & 0.130 & 0.181 & 0.215 & 0.033 & 0.040 & 0.045 & 0.064 \\
         & R & 0.216 & 0.216 & 0.187 & 0.187 & 0.113 & 0.113 & 0.050 & 0.050 & 0.019 & 0.023 & 0.157 & 0.157 & 0.135 & 0.135 & 0.163 & 0.163 & 0.038 & 0.038 & 0.056 & 0.068 \\
         & Hit Ratio & 0.216 & 0.216 & 0.187 & 0.187 & 0.113 & 0.113 & 0.050 & 0.050 & 0.019 & 0.023 & 0.157 & 0.157 & 0.135 & 0.135 & 0.163 & 0.163 & 0.038 & 0.038 & 0.056 & 0.068 \\

         \hline
         \multirow{6}{*}{\shortstack[c]{Relevance \\ Annotation \\ with LLM \\ (Mistral 7B)}} & MAP & 0.045 & 0.055 & 0.176 & 0.216 & 0.034 & 0.042 & 0.018 & 0.022 & -0.005 & -0.008 & 0.032 & 0.039 & 0.174 & 0.213 & 0.051 & 0.063 & 0.021 & 0.026 & -0.002 & -0.003 \\
         & MRR & 0.060 & 0.062 & 0.189 & 0.196 & 0.001 & 0.001 & -0.021 & -0.022 & -0.008 & -0.011 & 0.048 & 0.050 & 0.143 & 0.151 & 0.034 & 0.038 & -0.007 & -0.007 & 0.004 & 0.005 \\
         & NDCG & 0.049 & 0.060 & 0.178 & 0.218 & 0.032 & 0.039 & 0.018 & 0.022 & -0.006 & -0.009 & 0.036 & 0.044 & 0.175 & 0.214 & 0.049 & 0.060 & 0.022 & 0.028 & 0.000 & 0.000 \\

         & P & 0.028 & 0.034 & 0.137 & 0.166 & -0.004 & -0.006 & 0.021 & 0.025 & -0.005 & -0.008 & 0.002 & 0.003 & 0.138 & 0.167 & 0.010 & 0.013 & 0.014 & 0.017 & -0.006 & -0.010 \\

         & R & 0.014 & 0.014 & 0.032 & 0.032 & -0.016 & -0.016 & 0.019 & 0.019 & 0.003 & 0.003 & 0.000 & 0.000 & 0.039 & 0.039 & -0.042 & -0.042 & -0.017 & -0.017 & 0.017 & 0.021 \\

         & Hit Ratio & 0.014 & 0.014 & 0.032 & 0.032 & -0.016 & -0.016 & 0.019 & 0.019 & 0.003 & 0.003 & 0.000 & 0.000 & 0.039 & 0.039 & -0.042 & -0.042 & -0.017 & -0.017 & 0.017 & 0.021 \\

         
         \hline
         \multirow{6}{*}{\shortstack[c]{\metric \\ Annotations}} & MAP & 0.492 & 0.575 & 0.474 & 0.578 & 0.610 & 0.694 & 0.386 & 0.463 & - & - & 0.467 & 0.544 & 0.427 & 0.519 & 0.634 & \textbf{0.705} & 0.399 & 0.479 & - & - \\
         & MRR & 0.503 & 0.577 & \textbf{0.486} & 0.553 & \textbf{0.629} & 0.695 & \textbf{0.592} & \textbf{0.611} & - & - & 0.466 & 0.537 & 0.424 & 0.495 & \textbf{0.639} & 0.698 & \textbf{0.481} & \textbf{0.504} & - & - \\
         & NDCG & 0.505 & 0.590 & \textbf{0.486} & \textbf{0.592} & 0.612 & \textbf{0.697} & 0.404 & 0.484 & - & - & 0.481 & 0.560 & 0.440 & 0.536 & 0.635 & \textbf{0.705} & 0.403 & 0.484 & - & - \\
         & P\tabularnote{For non-integer relevance labels, precision is equal to average of the relevance labels.} & \textbf{0.529} & \textbf{0.598} & 0.484 & 0.577 & 0.594 & 0.663 & 0.329 & 0.391 & \textbf{0.504} & \textbf{0.669} & \textbf{0.522} & \textbf{0.586} & \textbf{0.482} & \textbf{0.571} & 0.633 & 0.695 & 0.378 & 0.449 & \textbf{0.540} & \textbf{0.712} \\
         & R & 0.519 & 0.519 & 0.426 & 0.426 & 0.619 & 0.619 & 0.301 & 0.301 & - & - & 0.488 & 0.488 & 0.408 & 0.408 & 0.631 & 0.631 & 0.299 & 0.299 & - & - \\
         & Hit Ratio\tabularnote{For non-integer relevance labels, hit ratio is equal to maximum of the relevance labels.} & 0.519 & 0.519 & 0.426 & 0.426 & 0.619 & 0.619 & 0.301 & 0.301 & 0.390 & 0.532 & 0.488 & 0.488 & 0.408 & 0.408 & 0.631 & 0.631 & 0.299 & 0.299 & 0.414 & 0.561 \\
         \hline
    \end{NiceTabular}}
    \label{tab:corr-bm25-contriever}
    \vspace{-0.2cm}
\end{table*}

\subsection{Setup}

\subsubsection*{\textbf{Datasets and Evaluation}}

We use Natural Questions (NQ) \cite{nq}, TriviaQA \cite{tqa}, HotpotQA \cite{hotpotqa}, FEVER \cite{fever}, and Wizard of Wikipedia (WoW) \cite{wow} datasets from the KILT \cite{kilt} benchmark. Due to the unavailability of ground truth labels for the test set, we utilize the publicly accessible validation set. As the retrieval corpus, we employ the Wikipedia dump of the KILT benchmark and adhere to the preprocessing outlined by \citet{dpr}, where each document is segmented into passages, each constrained to a maximum length of 100 words. The concatenation of the article title and passage is used as a document. The KILT benchmark furnishes document-level relevance labels (called Provenance) for its datasets, and these are employed for evaluating retrieval performance. In line with our preprocessing method, we define all passages within a positive document as positive passages for our evaluation. For relevance evaluation using an LLM, we employ Mistral\footnote{\url{https://huggingface.co/mistralai/Mistral-7B-Instruct-v0.2}} \cite{jiang2023mistral} to annotate each document within the retrieved list, determining whether it is relevant to the query or not. 
We adopt the metrics recommended by the KILT benchmark, namely Exact Match (EM) for NQ, TriviaQA, and HotpotQA, Accuracy for FEVER, and F1 for the WoW dataset. 

\subsubsection*{\textbf{Experiments Configuration}}

In all experiments, unless explicitly stated otherwise, we employ T5-small \cite{t5} with Fusion-in-Decoder (FiD) \cite{fid} as the LLM. We employ AdamW \cite{adamw} with a weight decay of $10^{-2}$ and a learning rate of $5 \times 10^{-5}$ for 10 epochs, incorporating linear warmup for the initial $5\%$ of training steps. The effective batch size is set to 64. Each model is trained using an A100 Nvidia GPU. For document retrieval during training, we utilize BM25 \cite{bm25} implemented in Pyserini \cite{10.1145/3404835.3463238} to retrieve 50 documents to augment the input with them. For fast vector search in dense retrieval with Contriever\footnote{\url{https://huggingface.co/facebook/contriever}} \cite{contriever}, we use Faiss \cite{douze2024faiss} flat index.

\subsection{Main Findings}


\subsubsection*{\textbf{How do different retrieval evaluation methods correlate with the end-to-end downstream performance in RAG?}}

To compare the different evaluation strategies for evaluating retriever in RAG, we report the correlation between the scores generated for each method and the downstream performance of the LLM (i.e., T5-small with FiD and 50 retrieved documents) in Table~\ref{tab:corr-bm25-contriever}. The results indicate that \metric attains the highest correlation compared to other evaluation approaches. Furthermore, the results show that regardless of the retrieval model employed, \metric consistently outperforms others in terms of correlation with the LLM's downstream performance. Interestingly, the most common approaches, KILT Provenance and Annotation with LLMs, that are, document-level relevance labels and using LLMs to assign a relevance label to each retrieved document, have the lowest correlation with the downstream performance of the LLM. This finding confirms that the LLM as the consumer of the retrieved results in RAG is the best judge for the performance of the retrieval model.

\subsubsection*{\textbf{How do different retrieval evaluation methods in RAG perform as the size of retrieval results increases?}}
\begin{figure}
    \centering
    \includegraphics[width=\linewidth]{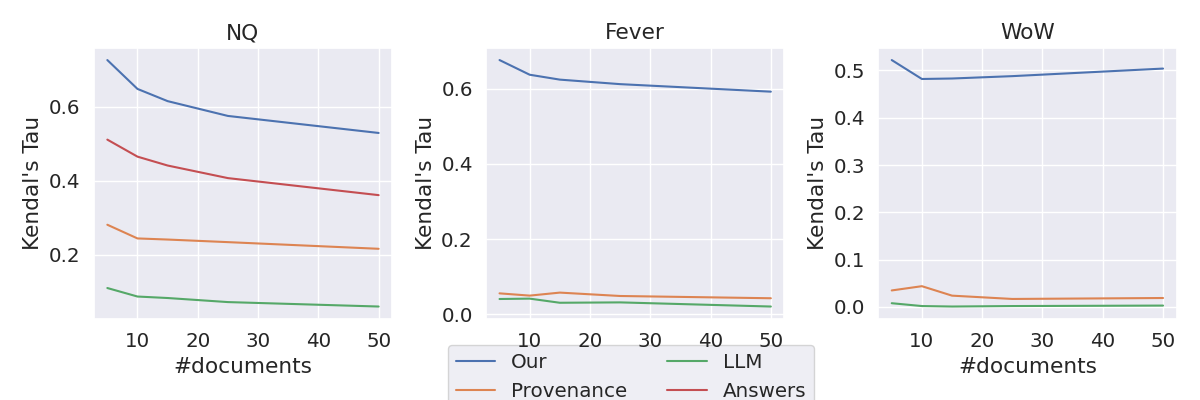}
    \vspace{-0.6cm}
    \caption{The correlation between evaluation approaches and the LLM's downstream performance varying number of retrieved documents by BM25. T5-small with FiD is used. The metric with the highest correlation in Table~\ref{tab:corr-bm25-contriever} is used.}
    \label{fig:ctx-size}
    \vspace{-0.5cm}
\end{figure}

To address this, we varied the number of retrieved documents and computed the correlation between the metric with highest correlation for each method in Table~\ref{tab:corr-bm25-contriever} at each specified number of retrieved documents and the downstream performance of the LLM given that number of retrieved documents. For the sake of space, we limit our experiments to three datasets: NQ for question answering, FEVER for fact-checking, and WoW for long-text generation. The results of this experiment are shown in Figure~\ref{fig:ctx-size}. The outcomes of this experiment reveal that irrespective of the quantity of retrieved documents, our suggested evaluation strategy consistently exhibits a higher correlation with the downstream performance of the LLM. Furthermore, the results illustrate that augmenting the number of retrieved documents leads to a decline in correlation—a intuitive observation, as all metrics assess each document-relevance label independently for scoring a ranked list, while the LLM uses information from the entirety of these documents to accomplish its task. 

\subsubsection*{\textbf{How does our method correlate with the downstream RAG performance as the size of large language models increases?}}

\begin{figure}
    \centering
    \includegraphics[width=\linewidth]{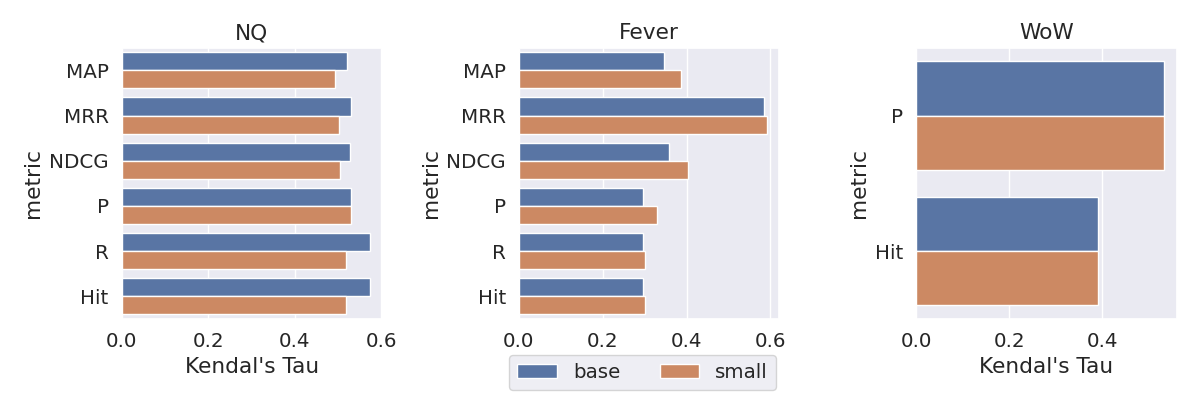}
    \vspace{-0.6cm}
    \caption{The correlation between \metric and the downstream performance of different LLM sizes. In this experiment, T5-small (60M parameters) and T5-base (220M parameters) with FiD are used. The documents are retrieved using BM25.}
    \label{fig:lm-size}
    \vspace{-0.4cm}
\end{figure}

In addressing this question, we computed the correlation between our retrieval evaluation strategy and the downstream performance of the LMs with two distinct sizes (i.e., T5-small with FiD consisting of 60M and T5-base with FiD  consisting of 220M parameters). For the sake of space, we limit our experiments to three datasets: NQ for question answering, FEVER for fact-checking, and WoW for long-text generation. The results illustrated in Figure~\ref{fig:lm-size} indicate that, for certain datasets, there is a higher correlation with the smaller LLM, while for others, a higher correlation is observed with the larger model. Nonetheless, in none of the cases is there a significant difference between the correlations, suggesting that the proposed approach is effective regardless of the LLM size.

\subsubsection*{\textbf{How does different retrieval-augmentation approaches affect the correlation between \metric and the downstream RAG performance?}}

\begin{figure}
    \centering
    \includegraphics[width=\linewidth]{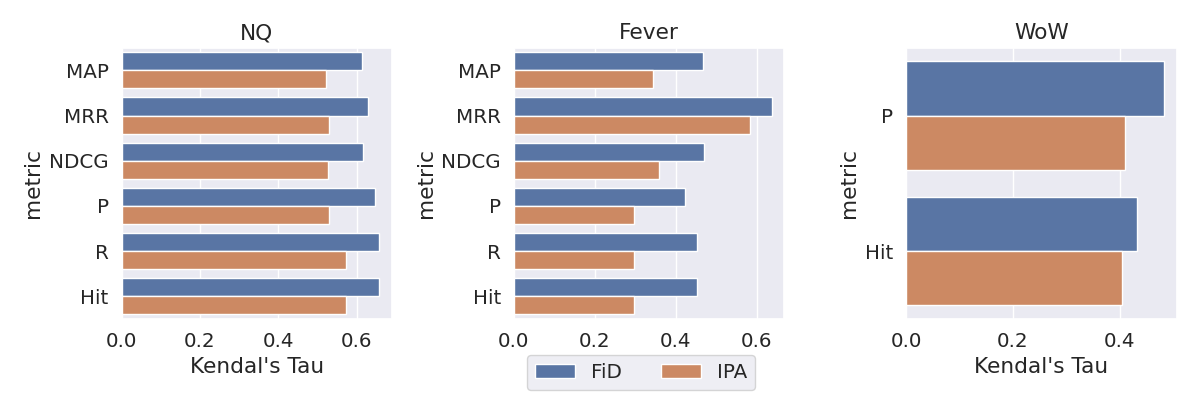}
    \vspace{-0.6cm}
    \caption{The correlation between \metric and the downstream performance of FiD and IPA LLMs. T5-small with 10 documents retrieved by BM25 is used. The number of documents is chosen based on the limitations of the input size in IPA.}
    \label{fig:ipa-vs-fid}
    \vspace{-0.4cm}
\end{figure}

We applied \metric to two LLMs. One LLM utilizes In-Prompt Augmentation (IPA), where the retrieved results are appended to the input of the LLM. The other LLM employs Fusion-in-Decoder (FiD) \cite{fid}, wherein each retrieved document is individually processed by the encoder, and subsequently, the representations for all documents are concatenated together and fed to the decoder. For the sake of space, we limit our experiments to NQ for question answering, FEVER for fact-checking, and WoW for long-text generation. The correlation between \metric and the outputs of each LLM is illustrated in Figure~\ref{fig:ipa-vs-fid}. Interestingly, the results suggest that although there is no significant difference between the correlation of \metric with IPA and FiD LLMs, \metric consistently exhibits a higher correlation with the FiD. This observation can be elucidated by considering the distinction between IPA and FiD methodologies. In IPA, all documents are concatenated together and then presented as a single input to the LLM. In contrast, FiD processes each document individually by feeding them separately to the LLM's encoder. Given that our approach aligns more closely with FiD, we believe this alignment is a contributing factor to the higher correlation between \metric and the downstream performance of FiD.

\subsubsection*{\textbf{How much more efficient is \metric compared to the end-to-end evaluation?}}

Here, we consider two factors: inference time and memory consumption. For inference time, we compare the total time required for end-to-end evaluation to generate scores with the total time used by \metric. In this experiment, we opt for the batch size of each approach to be as large as possible, maximizing the utilization of the entire GPU memory. The results of this experiment are reported in Table~\ref{tab:efficeincy}. The findings indicate that, on average, \metric is 2.468 times faster than end-to-end evaluation. Further elaborating, the speedup for \metric ranges from 1.232 to 3.252 times compared to end-to-end evaluation across the datasets, where the least speedup is for the long-text generation task (i.e., WoW).

To compare memory consumption between \metric and end-to-end evaluation, we conducted two experiments. First,  we compared the maximum memory required by end-to-end evaluation to assess a query with the maximum memory demanded by \metric for the same evaluation. To carry out this comparison, we configured the batch size for end-to-end evaluation to 1, while for \metric, we set it to the same number of documents used for one query by end-to-end evaluation (we call this query-level configuration). In the subsequent experiments, we set both batch sizes to 1 to assess the extent to which \metric demonstrates superior memory efficiency compared to end-to-end evaluation under the most efficient configuration (we call this document-level configuration). The results of these experiments are reported in Table~\ref{tab:efficeincy}. The findings indicate that in the query-level configuration, \metric exhibits between 7 to 15 times greater memory efficiency compared to end-to-end evaluation. Furthermore, in the document-level configuration, this efficiency gap widens, with \metric demonstrating 30 to 48 times more memory efficiency than end-to-end evaluation across different dataset. In summary, these experiments suggest that \metric is more efficient than end-to-end evaluation of a vanilla transformer, excelling in both inference time and memory utilization.

\begin{table}[]
    \centering
    \caption{The runtime and memory consumption of \metric in comparison with end-to-end evaluation. T5-small with FiD, consuming 50 documents is used.}
    \vspace{-0.4cm}
    \adjustbox{width=\linewidth}{
    \begin{tabular}{c|c|c|c|c|c}
         \multirow{2}{*}{Dataset} & \multicolumn{2}{c|}{Runtime (GPU)} & \multicolumn{3}{c}{Memory Consumption (GPU)} \\
         \cline{2-6}
         & E2E & \metric & E2E & \metric-Query & \metric-Document \\
         \hline
         NQ & 918 sec & 351 sec & 75.0 GB & 4.9 GB & 1.5 GB \\
         TriviaQA & 1819 sec & 686 sec & 46.2 GB & 5.4 GB & 1.5 GB \\
         HotpotQA & 1844 sec & 712 sec & 52.4 GB & 5.5 GB & 1.5 GB \\
         FEVER & 3395 sec & 1044 sec & 66.5 GB & 4.1 GB & 1.5 GB \\
         WoW & 912 sec & 740 sec & 47.9 GB & 6.5 GB & 1.5 GB \\
         \hline
    \end{tabular}}
    \label{tab:efficeincy}
    \vspace{-0.4cm}
\end{table}




\section{Conclusion}


This paper explores various approaches for evaluating retrieval models within a RAG pipeline. Additionally, it introduces \metric, a novel approach for evaluating retrieval models in the RAG pipeline. \metric leverages the per-document performance of the LLM on the downstream task to generate relevance labels. The findings suggest that the proposed approach exhibits significantly higher correlation with the downstream performance of the LLM. Furthermore, \metric demonstrates greater efficiency than end-to-end evaluation in terms of both memory consumption and inference time.

\section*{Acknowledgment}

This work was supported in part by the Center for Intelligent Information Retrieval, in part by Lowe’s, and in part by an Amazon Research Award, Fall 2022 CFP. Any opinions, findings and conclusions or recommendations expressed in this material are those of the authors and do not necessarily reflect those of the sponsor.

\bibliographystyle{ACM-Reference-Format}
\bibliography{XX-references}

\end{document}